\documentclass[sigconf]{acmart}
\usepackage{multirow}
\usepackage{makecell}
\usepackage{subfigure}
\usepackage{multicol}
\usepackage{multirow}
\usepackage{url}
\usepackage{amsmath,amsfonts,amsthm} 
\usepackage{graphicx}
\usepackage{subfigure}
 \usepackage{balance}
\usepackage{color}
\usepackage{tabularx}
\usepackage{algorithm}
\usepackage{algorithmic} 
\usepackage{stfloats}
\AtBeginDocument{%
  \providecommand\BibTeX{{%
    \normalfont B\kern-0.5em{\scshape i\kern-0.25em b}\kern-0.8em\TeX}}}

\setcopyright{acmcopyright}
\copyrightyear{2021}
\acmYear{2021}
\acmDOI{10.1145/1122445.1122456}

\acmConference[SIGIR '21]{SIGIR '21:In Proceedings of the International ACM SIGIR Conference
on Research and Development in Information Retrieval}{June 03--05, 2021}{Woodstock, NY}
\acmBooktitle{SIGIR '21:  ACM SIGIR Conference
on Research and Development in Information Retrieval,
  June 03--05, 2021, Woodstock, NY}
\acmPrice{15.00}
\acmISBN{978-1-4503-XXXX-X/18/06}

\begin{document}

\title{Data Distillation for Text Classification}


\author{Yongqi Li}\affiliation{The Hong Kong Polytechnic University\country{China}}\email{liyongqi0@gmail.com}
\author{Wenjie Li}\affiliation{The Hong Kong Polytechnic University\country{China}}\email{cswjli@comp.polyu.edu.hk}

\renewcommand{\shortauthors}{Anonymous Authors.}

\begin{abstract}
Deep learning techniques have achieved great success in many fields, while at the same time deep learning models are getting more complex and expensive to compute. It severely hinders the wide applications of these models. In order to alleviate this problem, {\it model distillation} emerges as an effective means to compress a large model into a smaller one without a significant drop in accuracy. In this paper, we study a related but orthogonal issue, {\it data distillation}, which aims to distill the knowledge from a large training dataset down to a smaller and synthetic one. It has the potential to address the large and growing neural network training problem based on the small dataset. We develop a novel data distillation method for text classification. We evaluate our method on eight benchmark datasets. The results that the distilled data with the size of 0.1\% of the original text data achieves approximately 90\% performance of the original is rather impressive. 
\end{abstract}



\maketitle

\section{Introduction}
Deep learning~\cite{lecun2015deep,hochreiter1997long} has achieved incredible success over the past years in a variety of applications ranging from computer vision~\cite{He_2016_CVPR} to natural language processing ~\cite{peng2018large}. To solve increasingly complex and difficult problems, deep learning models have shown a clear trend toward deeper and larger. The huge computational complexity and massive storage requirements post a big challenge to effectively train deep models by using massive training data. As everyone knows, the latest developed language model GPT-3 trained on 45 TB data contains about 175 billion parameters~\cite{brown2020language}, which makes it difficult to train, to fine-tune and even to use.

\begin{figure}
  \includegraphics[width=\linewidth]{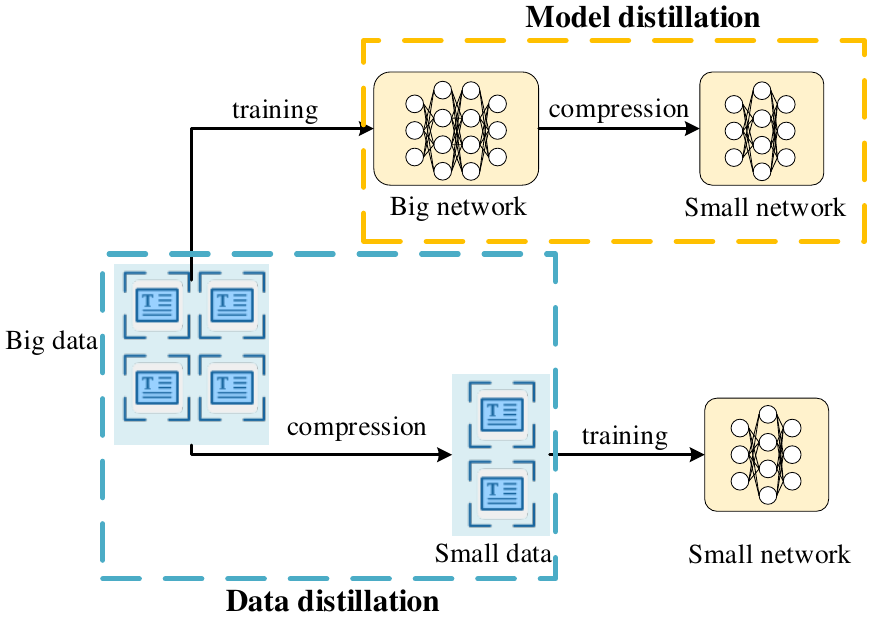}
  \vspace{-2em}
  \caption{Illustration of two flows to transfer the knowledge.}
  \label{example}
 \vspace{-2em}
\end{figure}

To provide efficient deep models for practical use, previous studies on {\it knowledge distillation}~\cite{hinton2015distilling} have attempted to compress a large and deep model down to a smaller one, which would not result in a significant drop in accuracy. We refer to the efforts along this line as {\it model distillation}. In model distillation, the student model mimics the teacher model to obtain a competitive or even a superior performance~\cite{gou2020knowledge,song2018neural}. Recently, another related but orthogonal task, {\it data distillation} starts to attract people’s attention. Different from model distillation that transfers knowledge from a large model to a small model, data distillation aims to encapsulate the knowledge of a large dataset into a small and synthetic dataset. The difference between them is visualized in Figure~\ref{example} above. On the one hand, to explore data distillation is of interest as a tool to study neural network generalization under small sample conditions. On the other hand, it has the potential to address the large and growing neural network training problem if the adequate neural networks can be quickly trained on the small distilled datasets rather than the original massive datasets. Moreover, data distillation helps to protect data privacy since the distilled dataset is a set of synthetic data, which will not disclose the original data.

As a matter of fact, besides data distillation, a plenty of methods are targeted to reduce the size of datasets for different purposes. For instance, active learning aims to reduce the required size of training data by labeling only the examples that are determined to be the most important~\cite{spina2015active,cohn1996active}. In nearest-neighbor classification, prototype selection ~\cite{garcia2012prototype} are also investigated to improve classification efficiency. In general, the above-mentioned methods attempt to select samples from the true distributions, i.e., the subsets of the original training sets. Differently, data distillation seeks to {\it generate} not to {\it select} a small dataset that contains most of the knowledge of the original dataset. Selecting a subset of the original dataset has a clear upper bound, because it always happens that some knowledge of the deserted data is not contained in the selected dataset. However, there is a possibility to generate new data to cover all the original knowledge. Please also note that, compared with another popular generation technique based on generative adversarial networks~\cite{NIPS2014_5ca3e9b1}, data distillation mainly focuses on knowledge transfer from a large dataset to a small dataset rather than creating lifelike samples. The challenge in developing effective data distillation solutions is distinctive. 

\begin{figure*}[t]
  \includegraphics[width=1.0\linewidth]{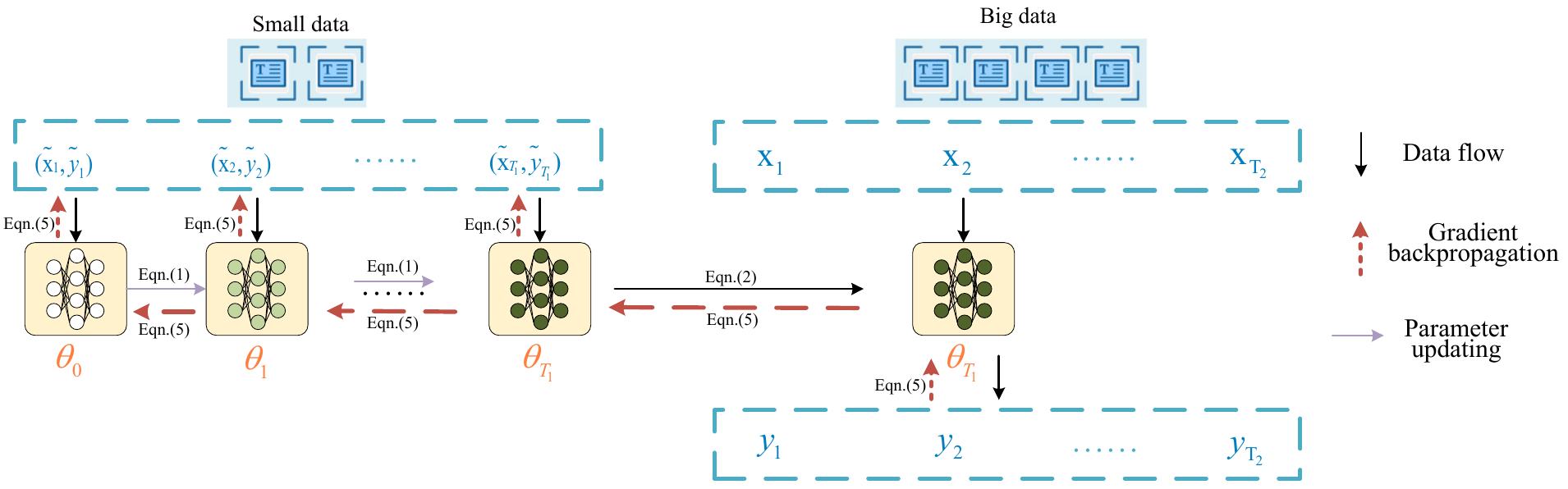}
  \vspace{-1em}
  \caption{Illustration of our proposed text data distillation method.}

  \label{framework}
\end{figure*}

In this paper, we propose a viable data distillation method for text classification, with the aim of distilling a large class labelled dataset into a small one without a significant drop in classification accuracy. For each class, we randomly initialize a handful of samples in the form of numeric matrixes, called {\it distilled data}. We then design an optimization scheme to update the numeric distilled data towards the direction of the original data. It is of particular concern to us how to build the bridge to connect the two training processes that use the distilled data and the original data, respectively. We come up with the following idea. When we use the distilled data to train the model, the network weights are derived from a differentiable function taking the synthetic distilled data as the dependent variable by design. Given this function, the gradients can be backpropagated to the distilled data when using the original data to train the network, as illustrated in Figure~\ref{framework}. By repeating the above processes alternately, the distilled data is continuously updated to better approximate the original data. Finally, the distilled synthetic data can be used to train any other model as the normal training data. 


We evaluate our method on eight benchmark datasets. The results are quite encouraging. The distilled data with the size of 0.1\% of the original data actually achieves approximate 90\% performance of the original and it significantly outperforms the heuristically selected examples with the same size.

\section{Approach}
Let the original text classification dataset denoted as $\mathcal{D}=\{x_i, y_i\}_{i=1}^N$, where $x_i$ is a piece of text, $y_i$ is its class label, and $N$ is the number of samples in $\mathcal{D}$. We expect to transfer the knowledge in $\mathcal{D}$ down to a new, much-reduced synthetic dataset $\tilde{\mathcal{D}}=\{\tilde{x}_i, \tilde{y}_i\}_{i=1}^M$, where $M$ is the number of samples in $\tilde{\mathcal{D}}$ and $M<<N$. 

The discrete nature of text data makes data distillation a challenging problem. We choose to generate human-unreadable numeric matrixes instead of extracting real text words to form the distilled data, considering the target is to generate data for neural networks to learn knowledge from it  rather than for people to read and understand. More important, the numeric matrixes can be dealt with in the same way as the parameters of a network, thus a gradient descent based method can be applied to update and optimize it. Even though, it is non-trivial to develop a suitable framework and to design an appropriate training objective that can effectively update the distilled data towards the original data. Inspired by the previous works \cite{wang2018dataset,bohdal2020flexible}, we made efforts to conduct a function where the input is the distilled data and output is a well-trained model, so that the gradients can be backpropagated to the distilled data via this function when using the original data to train the model. 

In more details, we first randomly initialize the distilled data $\tilde{\mathcal{D}}$ as $\tilde{\mathcal{D}}_0$. For each class, we create a handful of matrixes. Then we apply minibatch stochastic gradient descent to train a text classification model $\Theta_0$ on the initial distilled data. Specifically, we divide the distilled data $\tilde{\mathcal{D}}_0$ into batches, denoted as $\{\tilde{x}_t\}_{t=1}^{T_1}$, where $T_1$ is the number of batches. We input the batches of data to the model to update its parameters. Given a batch of data $\tilde{x}_t$, the parameters are updated as follows,
\begin{equation} \label{eqn1}
\Theta_t=\Theta_{t-1}-\alpha*\partial_{\Theta_{t-1}} L(\tilde{x}_t,\Theta_{t-1}),
\end{equation}
where $\alpha$ denotes the learning rate and $L()$ denotes the loss function. We refer to the model trained after the $T_1$ batches as $\Theta_{T_1}$\footnote{In practice, the training process can be extended to multi epochs.}. Basically, $\Theta_{T_1}$ is derived from $\Theta_0$ via a series of gradient descent steps by,
\begin{equation} \label{eqn2}
\Theta_{T_1}=F(\Theta_0, \alpha, \tilde{\mathcal{D}}),
\end{equation}
where $F()$ denotes the training process. Note that $F(\Theta_0, \alpha, \tilde{\mathcal{D}})$ is a differentable function on its independent variable.

We then transform the text $x_i$ in the original text data $\mathcal{D}$ to an embedding matrix. By convention, we pad $x_i$ to a fixed length and embed each word in $x_i$ to an embedding vector. Similarly, we divide the data into batches, denoted as $\{x_t\}_{t=1}^{T_2}$. We input a batch of data $x_t$ to the model $\Theta_t$ and calculate the loss $L(x_t,\Theta_{T_1})$ as follows,
\begin{equation} \label{eqn3}
L(x_t,\Theta_{T_1})=L(x_t,F(\Theta_0, \alpha, \tilde{\mathcal{D}})).
\end{equation}

As mentioned before, $F(\Theta_0, \alpha, \tilde{\mathcal{D}})$ is differentiable, thus the current loss function $L(x_t,\Theta_{T_1})$ is also a differentiable function of the distilled data $\tilde{\mathcal{D}}$. Our training objective is to find out an optimal $\tilde{\mathcal{D}}_*$ that minimizes the loss, formulated as,
\begin{equation} \label{eqn4}
\tilde{\mathcal{D}}_*=arg \ \underset{\tilde{\mathcal{D}}}{min} \ L(x_t,\Theta_{T})=arg \ \underset{\tilde{\mathcal{D}}}{min} \ L(x_t,F(\Theta_0, \alpha, \tilde{\mathcal{D}})).
\end{equation}

Since the distilled data $\tilde{\mathcal{D}}$ is numeric like the parameters of a network, $L(x_t,\Theta_{T})$ is a differentiable function of the synthetic distilled data according to Equation ($\ref{eqn3}$). It makes workable to update the numeric $\tilde{\mathcal{D}}$ using the following gradient descent algorithm,
\begin{equation} \label{eqn5}
\tilde{\mathcal{D}}_t=\tilde{\mathcal{D}_{t-1}}-\alpha*\partial_{\tilde{\mathcal{D}}} L(x_t,\Theta_{T_1}).
\end{equation}

Through the above training steps, the distilled data $\tilde{\mathcal{D}}$ is updated towards the minimal loss, which means that the difference of the results between using $\tilde{\mathcal{D}}$ and ${\mathcal{D}}$ is reduced. Using the batches of data $\{x_t\}_{t=1}^{T_2}$, the distilled data is updated from $\tilde{\mathcal{D}}_0$ to $\tilde{\mathcal{D}}_{T_2}$ via Equation. ($\ref{eqn5}$). At the end, we store the distilled data $\tilde{\mathcal{D}}_{T_2}$ and it can be used to train neural networks as normal text data.

In short, the number of the samples in $\tilde{\mathcal{D}}_{T_2}$ is $M$, which is much smaller than that of the original data $\mathcal{D}$. At the same time, the distilled data $\tilde{\mathcal{D}}_{T_2}$ contains knowledge of the original data as much as possible according to the training objective in Equation ($\ref{eqn4}$) and through the training process illustrated in Equation ($\ref{eqn5}$). To evaluate the proposed method, we can train an arbitrary text classification model on both the distilled data and the original data, and compare their performance on a same test data.

\section{Experiments}
\subsection{Datasets}
We used eight publicly benchmark datasets~\cite{zhang2015character,qiao2018new,du2019explicit} to evaluate our proposed text data distillation method. These datasets are sourced from various tasks, including sentiment analysis, news classification, question answering and ontology extraction. The statistics of the datasets are summarized in Table~\ref{dataset statistics}. We applied our method to the training set of the datasets and obtained the much-reduced distilled training data.

\begin{table}[tbp]
\centering
\caption{Statistics of 8 Datasets.}\label{dataset statistics}
 \vspace{-1em}
 \scalebox{0.9}{
\begin{tabular}{ccccc}
\toprule
\multirow{2}{*}{Dataset}&\multirow{2}{*}{Classes}&Train&Test&\multirow{2}{*}{Task}\\
&&Samples&Samples&\\

\midrule
\multirow{2}{*}{DBpedia}&\multirow{2}{*}{14}&\multirow{2}{*}{560k}&\multirow{2}{*}{70k}&Ontology\\
&&&&Extraction\\
\midrule
Yahoo! Answers&10&1400k&60k&QA\\
\midrule
Sogou News&5&450k&60k&\multirow{1}{*}{News}\\
AG' News&4&120k&7.6k& Classification\\

\midrule
Yelp Review Full&5&650k&50k&\\
Yelp Review Polarity&2&560k&38k&Sentiment\\
Amazon Review Full &5&3,600k&400k&Analysis\\
Amazon Review Polarity&2&3,000k&650k&\\
\bottomrule
\end{tabular}}
\end{table}

\subsection{Experimental Settings}
\textbf{Evaluation Protocols}. To evaluate the utility of the distilled data, we train a text classification model with the same network structure on the original data and our distilled data, respectively. And then we calculate the accuracy of the two well trained model on the same test set. We also compare the distilled data with the same size of the data randomly selected from the original data.

\textbf{Implement Details}. 
We use the pre-trained GloVe word vectors\footnote{\url{https://nlp.stanford.edu/projects/glove/.}}, which is trained on the Twitter data and the dimension is 100 for all of the datasets except for the Sogou News dataset. Because there are too many words out of the vocabulary in the Sogou News dataset, we randomly initialized the word vectors. We follow the work~\cite{zhang2015character} that released the eight benchmark datasets to apply the TextCNN network. We do not use any extra regularization method, like L2 normalization or dropout. Note that Equation ($\ref{eqn5}$) involves high order derivative, which requires expensive memory and extensive computation. We therefore apply the back-gradient optimization technique that formulates the necessary second order terms into efficient Hessian-vector products~\cite{pearlmutter1994fast} so that they can be can be easily calculated with modern automatic differentiation systems such as PyTorch~\cite{paszke2019pytorch}.

\subsection{Experiment Results}
The test set accuracies of the text classification models trained on the full training data, the randomly selected data, and the distilled data are summarized in Table~\ref{overall}. The numbers of samples in the random data and the distilled data are the same, i.e., 0.1\% of the original training data in DBpedia, yahoo! Answers, Sogou News, AG’ News, Yelp Review Full, and Yelp Review Polarity datasets. Considering the larger scale of Amazon Review Full and Amazon Review Polarity, we set the size to 0.01\% of the original training data on these two datasets. The findings from this set of experiments are as follows. 

(1) It is observed that the text classification model trained on the full data obtains the best performance. This is not surprising, since the sizes of the random data and distilled the data are much smaller than that of the full data. Moreover, the random data is only a part of the full data and the distilled data is updated towards the full data. Therefore, the full data can be regarded as the upper bound. 

(2) Although the size of the distilled data is much smaller than the full data, the model trained on it can still achieve a comparative performance. In terms of accuracy, the distilled data obtains 81.78\%, 98.35\%, 92.06\%, 97.11\%, 94.00\%, 94.25\%, 83.47\%, and 93.76\% relative to the full data (referred to as the upper bound) on the DBpedia, yahoo! Answers, Sogou News, AG’ News, Yelp Review Full, Yelp Review Polarity, Amazon Review Full, and Amazon Review Polarity, respectively. In average, up to 91.84\% of the accuracy trained on the full data is achieved. This impressive result verifies the effectiveness of our proposed method and demonstrates the great potential of \textit{data distillation}. It is worth mentioning that on the two Amazon Review datasets, the distilled data only covers 0.01\% of the full training data, while it is 0.1\% on the other datasets. Therefore, the gap between the distilled data and full data on these two datasets is a big larger than that on the other datasets. 

(3) The distilled data significantly surpasses the random data. As claimed before, generating synthetic samples is more possible to get close to the upper bound because it aims to compresses all knowledge into a small data rather than select a subset. We also find that although the samples in the random data only account for a small percentage of the full data, it also supports to train a decent classification model successfully in some datasets. For example, for each class on the AG's News dataset there are only 30 samples. The classification model trained on the small random data still achieves 0.7133 accuracy. It is demonstrated that there are many redundancies and repetitive knowledge, and further shows the necessary of data distillation.

\begin{table*}[tp]
\centering
\caption{Accuracy on 8 benchmark datasets. Full data and Random data denote the original and randomly selecting a subset of the training data, respectively. The size of the Random data and our Distilled data is same. And the size is 0.01\% of the Full data on Amazon Review Full, Polarity datasets, while it is 0.1\% on other datasets. }\label{overall}
\vspace{-0.5em}
 \scalebox{1.0}{
\begin{tabular}{ccccccccc}
\toprule
\multirow{2}{*}{Method}&\multirow{2}{*}{DBpedia}&Yahoo!  &Sogou& \multirow{2}{*}{AG’s News}&Yelp &Yelp& Amazon & Amazon\\
&&Answers&News&&Review Full&Review Polarity&Review Full&Review Polarity\\
\midrule
Full data&0.9779&0.6700&0.9404&0.8819&0.6045&0.9219&0.5464&0.9102\\
Random data&0.6910&0.5293&0.8278&0.7133&0.3582&0.7763&0.2603&0.6952\\
Distilled data&0.7998&0.6590&0.8658&0.8564&0.5682&0.8689&0.4561&0.8534\\
\bottomrule
\end{tabular}}
\end{table*}

\begin{figure}[tbp]
  \centering
  \subfigure[ ]{
  \includegraphics[width=0.22\linewidth]{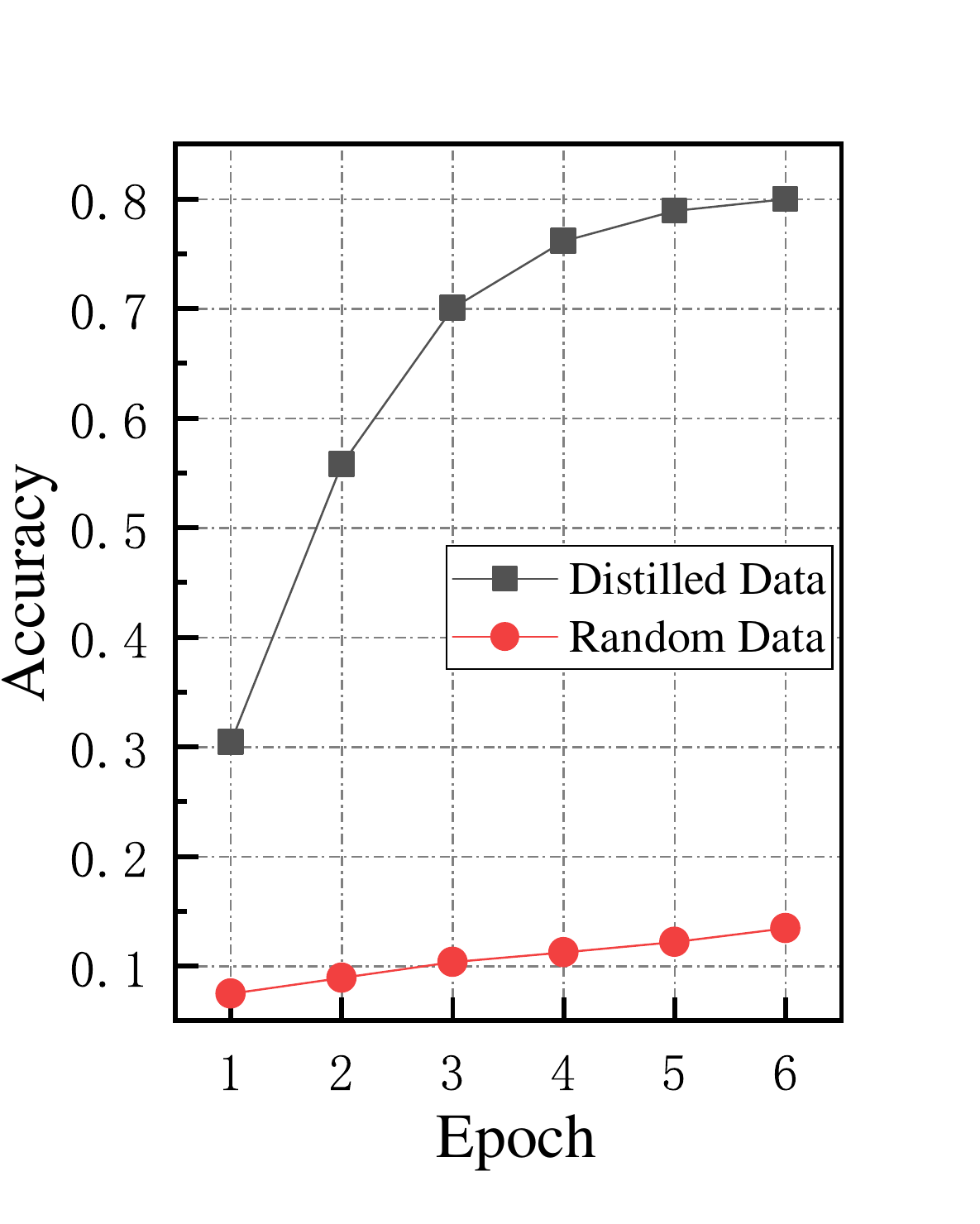}
   \label{fig:subfig1}
   }
     \subfigure[]{
  \includegraphics[width=0.22\linewidth]{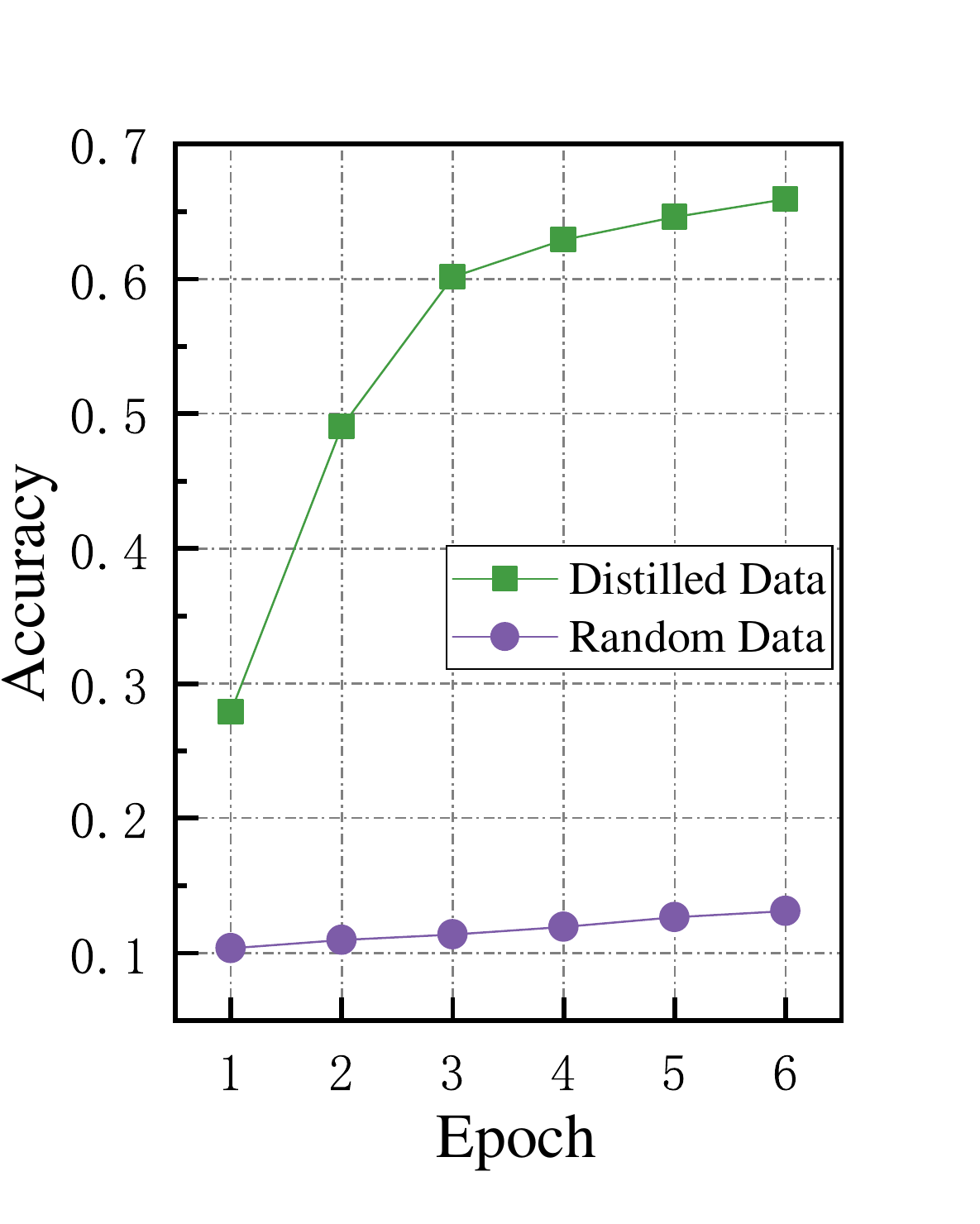}
   \label{fig:subfig2}
   }
        \subfigure[ ]{
  \includegraphics[width=0.22\linewidth]{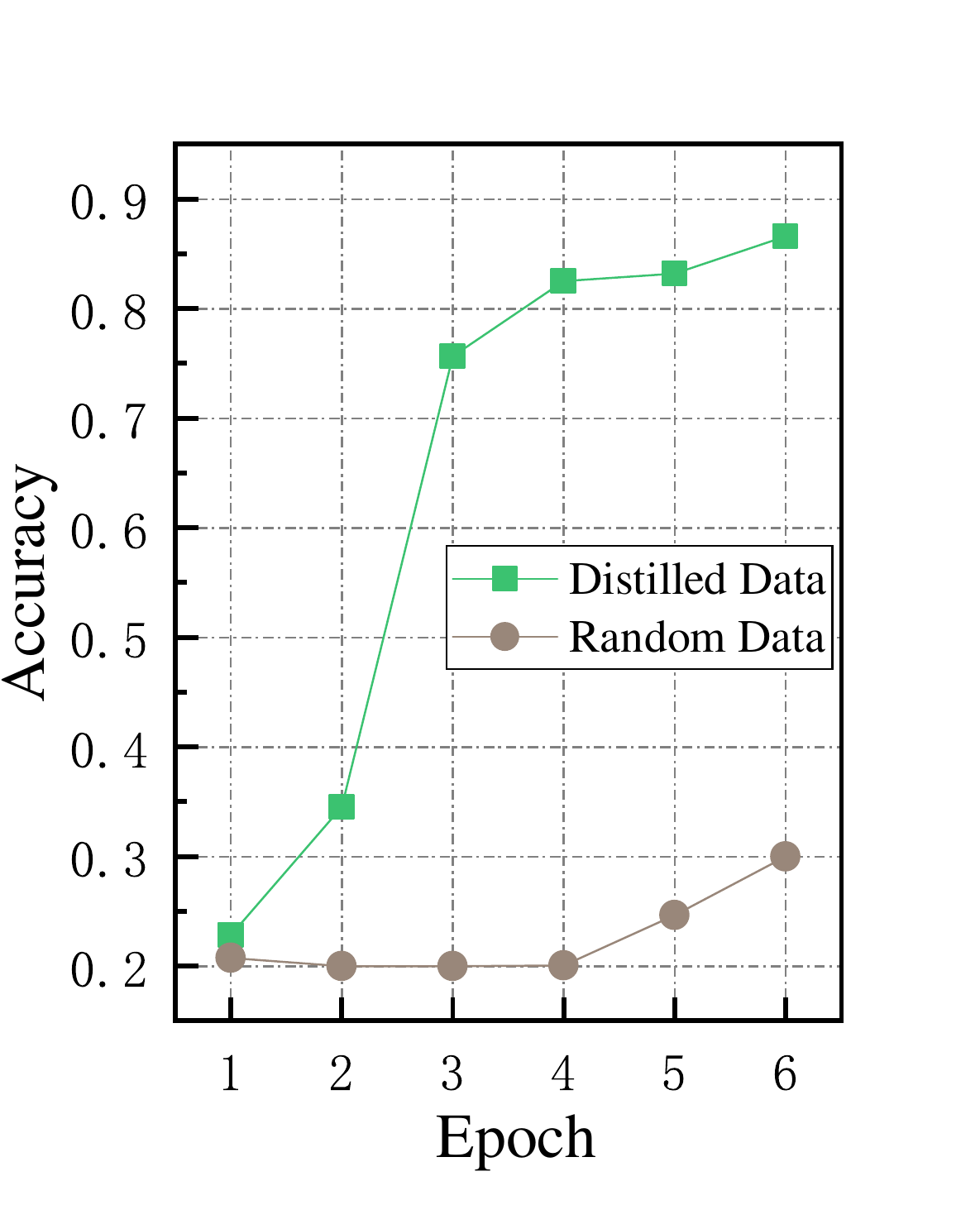}
   \label{fig:subfig3}
   }
        \subfigure[ ]{
  \includegraphics[width=0.22\linewidth]{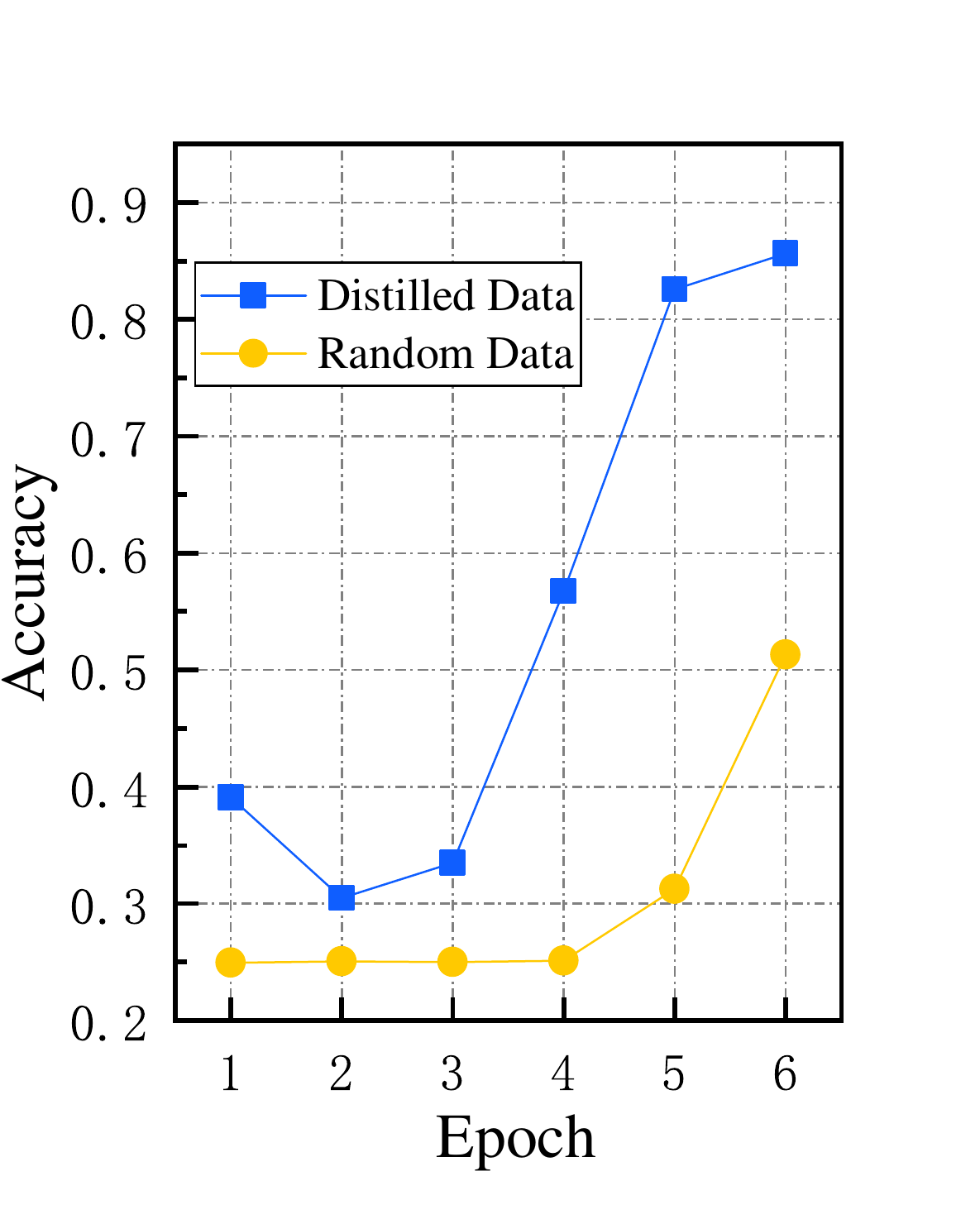}
   \label{fig:subfig4}
   }
        \subfigure[ ]{
  \includegraphics[width=0.22\linewidth]{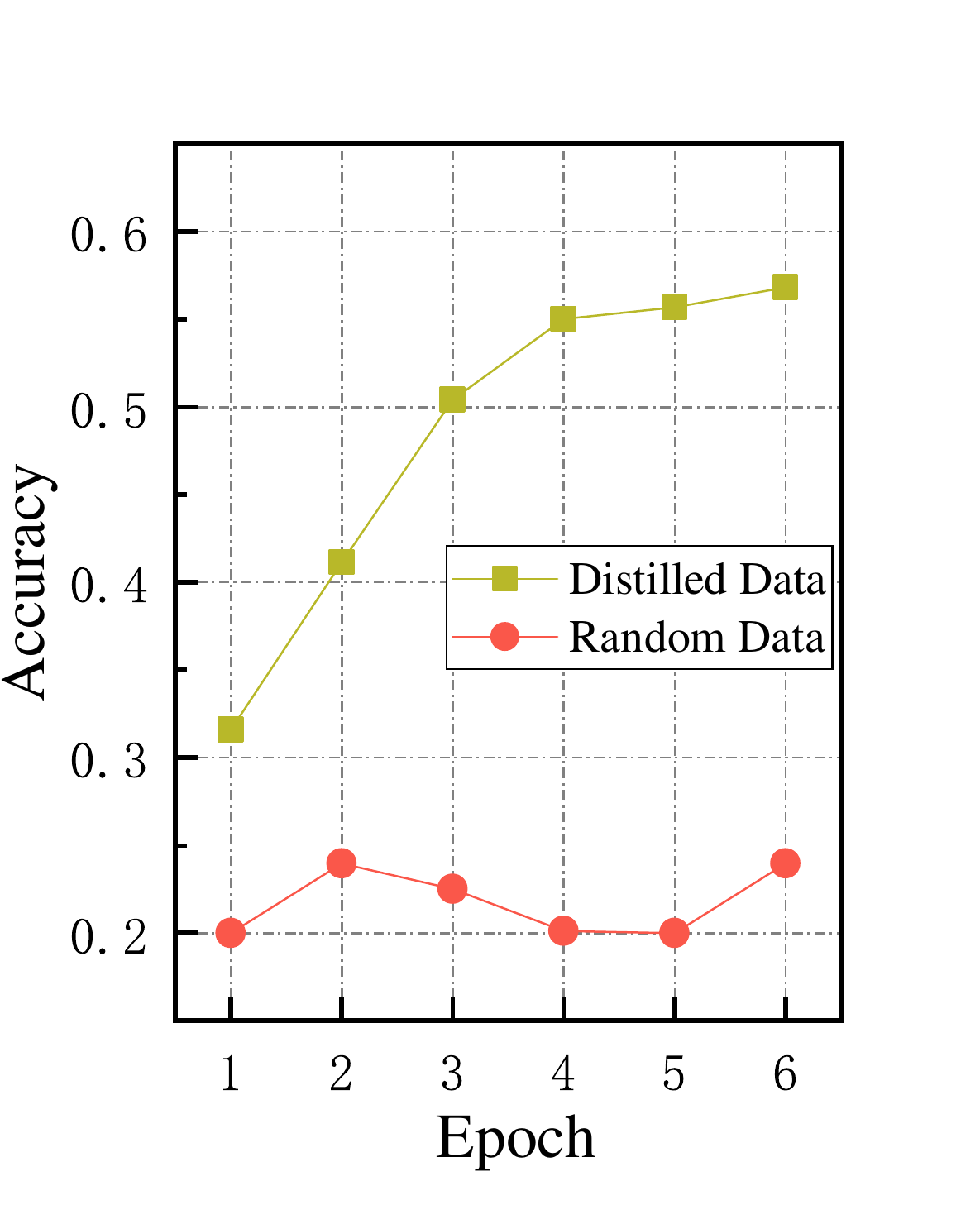}
   \label{fig:subfig5}
   }
        \subfigure[ ]{
  \includegraphics[width=0.22\linewidth]{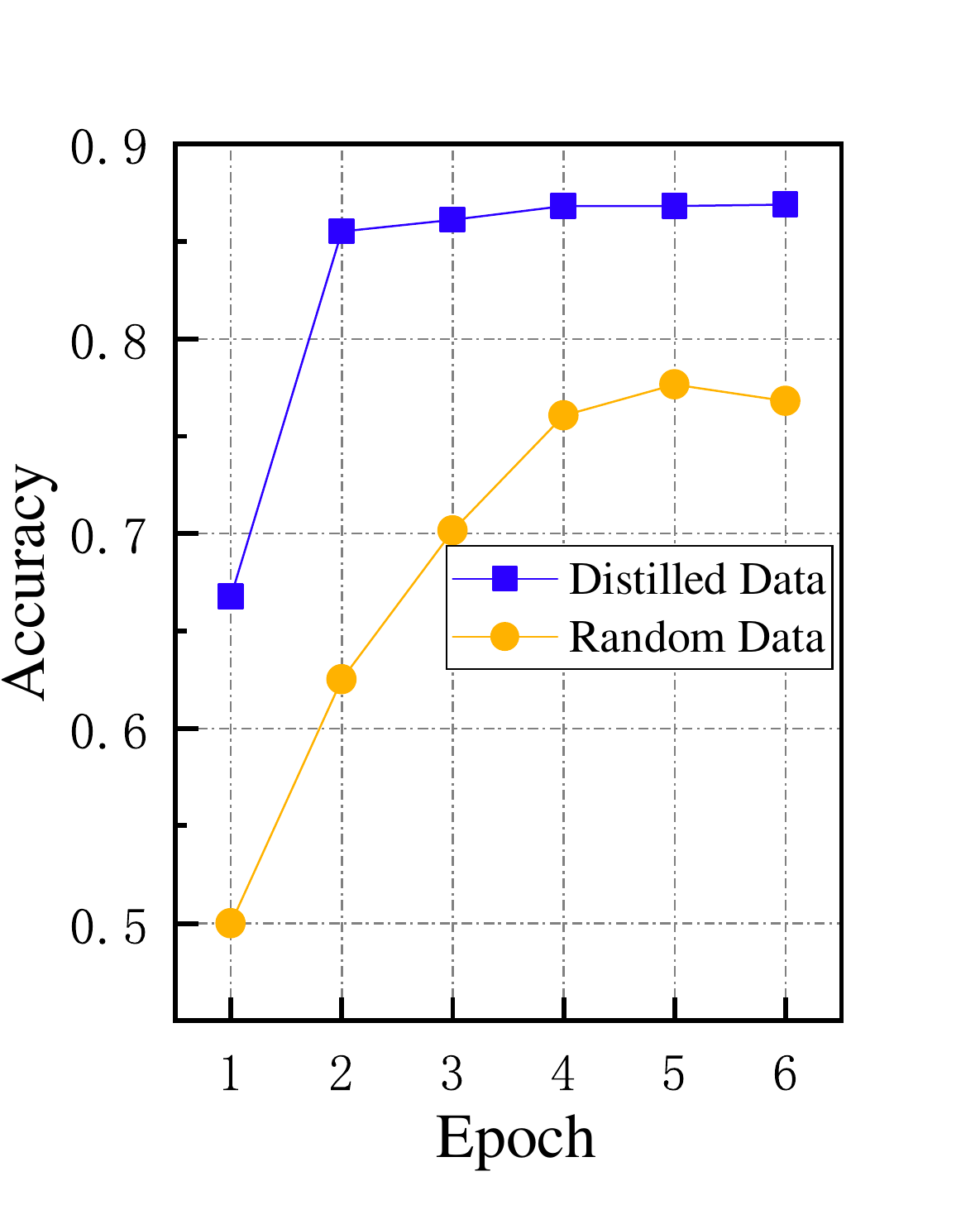}
   \label{fig:subfig6}
   }
        \subfigure[ ]{
  \includegraphics[width=0.22\linewidth]{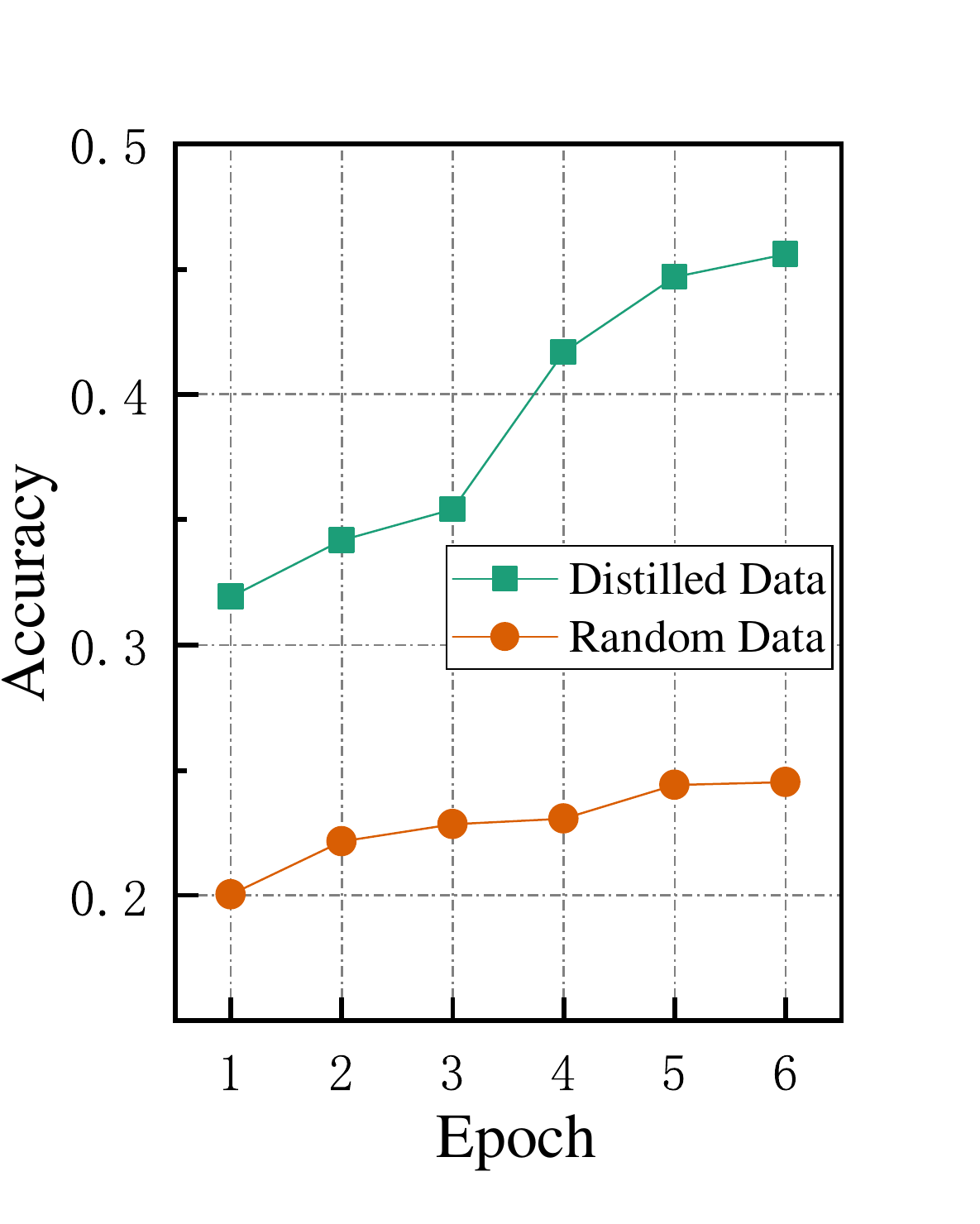}
   \label{fig:subfig7}
   }
        \subfigure[ ]{
  \includegraphics[width=0.22\linewidth]{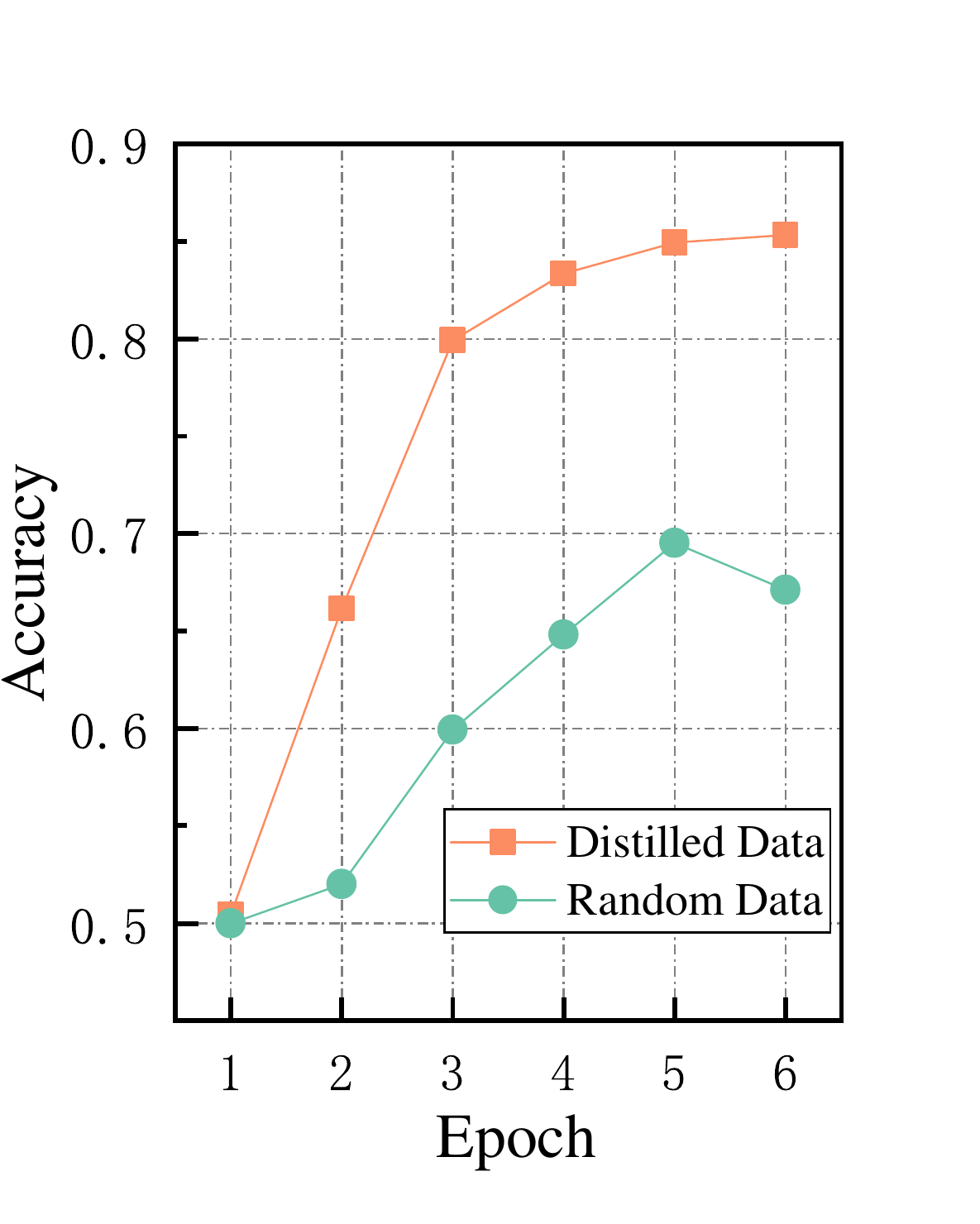}
   \label{fig:subfig8}
   }

\vspace{-1.5em}
  \caption{Accuracy versus the training epochs. (a)-(h) performance over DBpedia, Yahoo! Answers, Sogou News, AG’ News, Yelp Review Full, Yelp Review Polarity, Amazon Review Full, and Amazon Review Polarity respectively. }\label{training_speed}
\vspace{-1.5em}
\end{figure}

We further compare the performance of the model trained on the random data and the distilled data versus the training epochs. By analyzing Figure~\ref{training_speed}, we gain the following insights.

(1) It is easier to train neural networks on the distilled data compared with the random data. For example, in Figures~\ref{fig:subfig1} and~\ref{fig:subfig1}, the accuracy of the model trained on our distilled data increases fast to reaches its highest level. This demonstrates that the distilled data facilitate the effective training process, which is also verified in~\cite{wang2018dataset}. It is contributed to our optimization scheme that generates a more smooth optimization space when distilling knowledge.

(2) The accuracy of the model trained on the random data gradually ascends but still has a big gap to reach its best performance as reported in Table~\ref{overall}. For example, the accuracy of the model trained on the random data after 6 training epochs is 0.1345 in Figure~\ref{fig:subfig1}, and  more training epochs are needed in order to reach its best accuracy 0.6910. It is understandable that the model trained on the random data requires more training epochs to learn knowledge, since there are only a few numbers of samples for each class. We also find that in Figure~\ref{fig:subfig6}, the model trained on the random data converges as fast as on the distilled data. This might because there are only two classes, thus it is easier for a classification model to distinguish samples.

\begin{figure}[tbp]
  \centering
  \subfigure[Sogou News]{
  \includegraphics[width=0.44\linewidth]{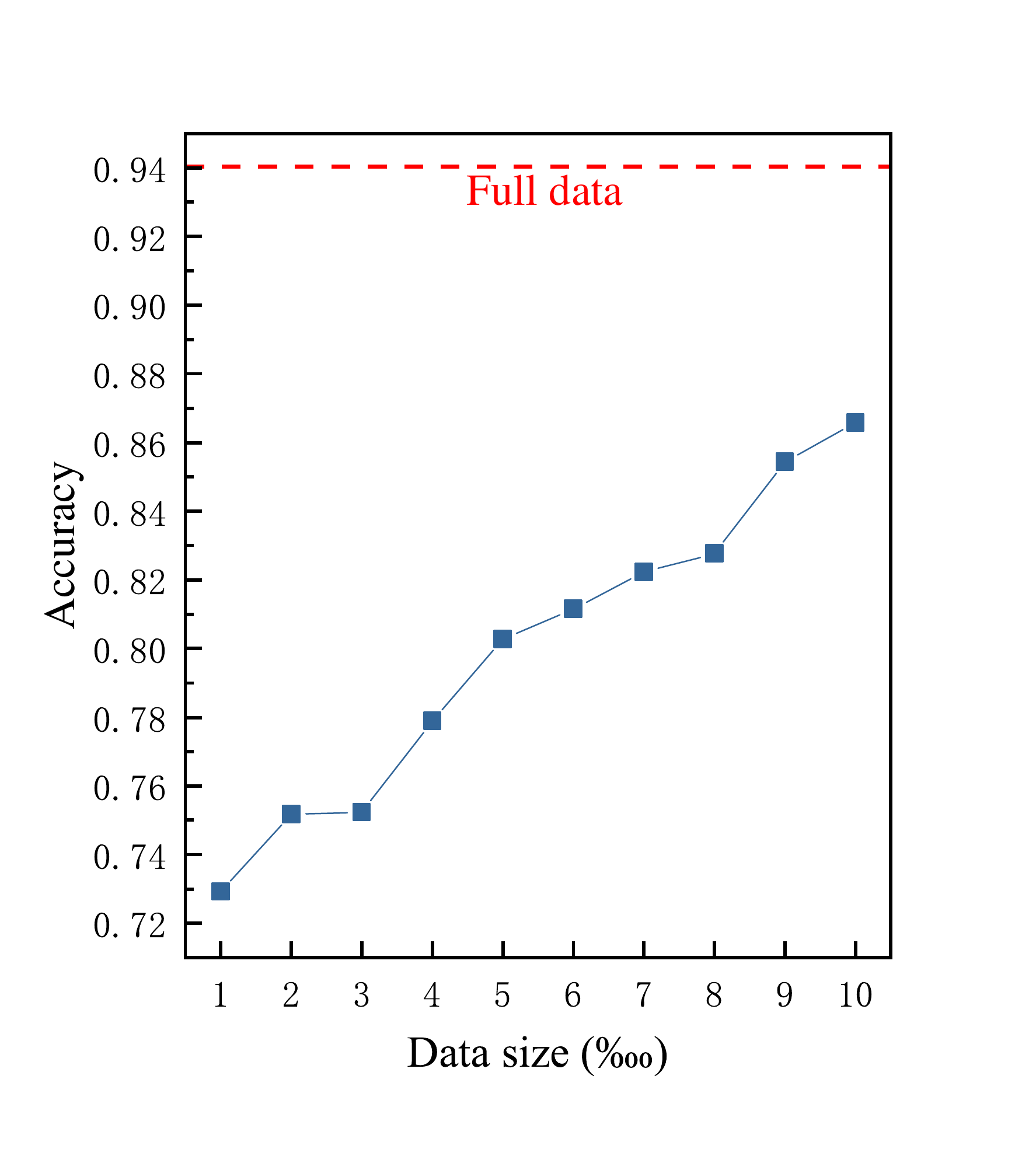}
   \label{fig4:subfig1}
   }
     \subfigure[AG's News]{
  \includegraphics[width=0.44\linewidth]{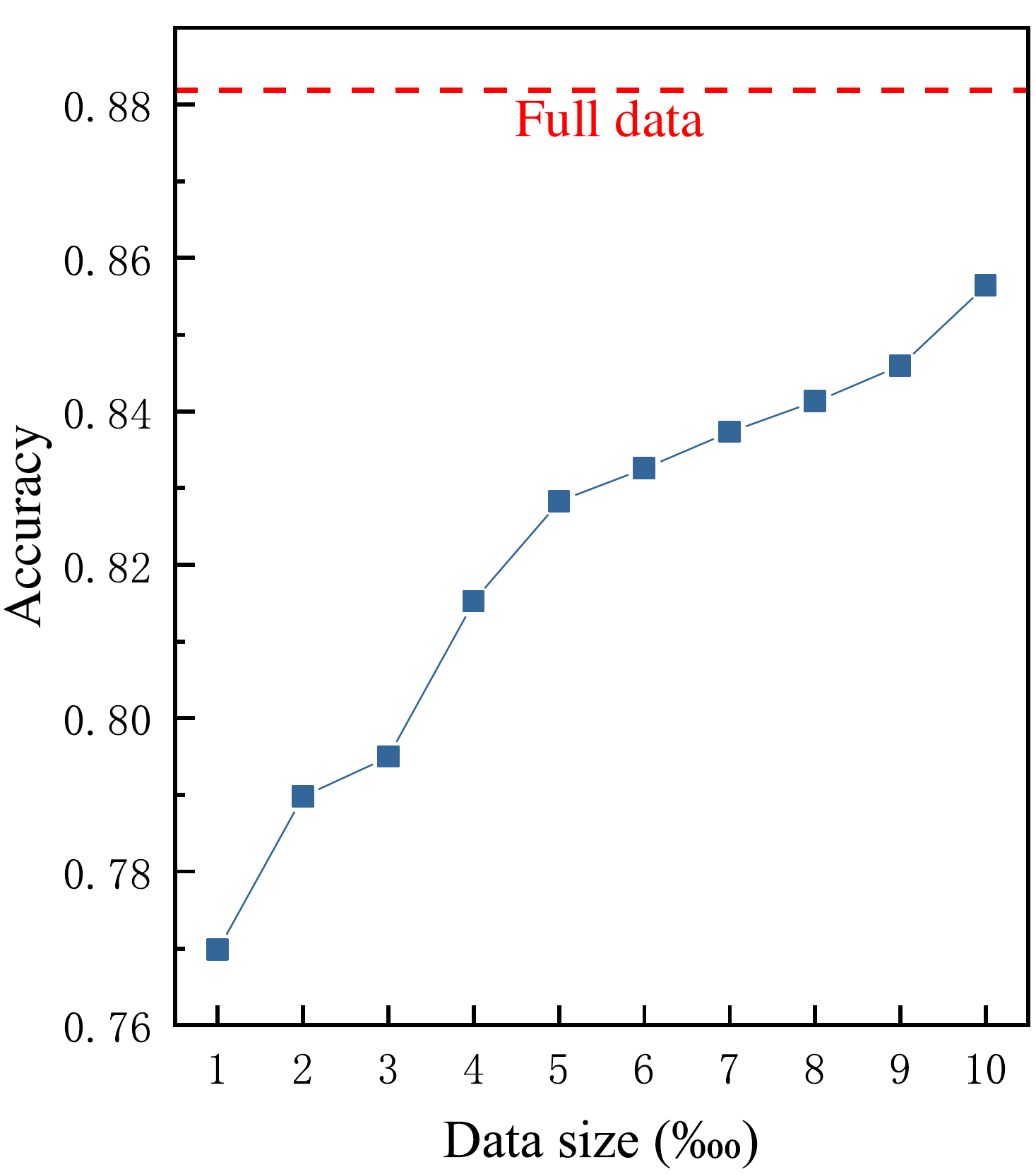}
   \label{fig4:subfig2}
   }

\vspace{-1.5em}
  \caption{Accuracy versus the size of the distilled data.}\label{data_size}
\vspace{-1.5em}
\end{figure}

\subsection{Data Size Analysis}
In addition to compare with the full data and the random data, we also conduct experiments to explore the influence of the distilled data size on the performance. Towards this end, we report the accuracy with the data size varying from 0.01\% to 0.1\% of the full data, as shown in Figure~\ref{data_size}. It is observed that when the size of the distilled data increases, the accuracy of the model trained on it raises and gets closer to the upper bound, i.e., the model trained on the full data. It may be due to the fact that it is much likely for the larger size of data to cover the knowledge distilled from the full data. Although we formulate the necessary second order terms into efficient Hessian-vector products as mentioned before, the computing consumption is still large. Thus, we only extend the size of the distilled data to 0.1\% of the full data. This is the limitation of the current data distillation method and requires further improvement.

\section{Conclusion}
In this paper, we explore a novel problem, data distillation, which aims to distill knowledge from a large training dataset down to a smaller and synthetic one. We propose a viable data distillation method for text classification. An optimization scheme is designed to update the numeric distilled data towards the direction of the original data. The experimental result that the distilled data with the size of 0.1\% of the original text data achieves approximately 90\% performance of the original is rather impressive. It also shows that the small distilled data facilitates the effective training process.

\bibliographystyle{ACM-Reference-Format}

\bibliography{acmart}

\end{document}